\documentclass[letterpaper, 10 pt, conference]{ieeeconf}  

\usepackage{mathtools}
\usepackage{amsmath}
\usepackage{esint}
\usepackage{mathdesign}

\usepackage{algorithm}
\usepackage{algorithmicx}
\usepackage{algpseudocode}

\usepackage{booktabs}
\usepackage{tabularx}

\usepackage{graphicx}
\usepackage{subfig}
\usepackage{bm}
\usepackage[ labelsep= period]{caption}

\usepackage{bm}
\usepackage[ labelsep= period]{caption}
\usepackage{color}

\def \rrr{\color{red}}
\def \kkk{\color{black}}

\IEEEoverridecommandlockouts                              

\title{\LARGE \bf
	Real-Time Salient Closed Boundary Tracking via Line Segments Perceptual Grouping
}

\author{Xuebin Qin$^\star$, Shida He, Camilo Perez Quintero, Abhineet Singh, Masood Dehghan and Martin Jagersand 
	\thanks{ The authors are with the Dept. of Computing Science, University of Alberta, Canada. 
{\tt\small \{xuebin, shida3, caperez, asingh1, masood1, mj7\}@ualberta.ca}. } 
\thanks{ Xuebin  was supported by China Scholarship Council and University of Alberta.}
}

\begin{document}
	
	\maketitle
	\thispagestyle{empty}
	\pagestyle{empty}

\begin{abstract}
		
  This paper presents a novel real-time method for tracking salient closed boundaries from video image sequences. This method operates on a set of straight line segments that are produced by line detection. The tracking scheme is coherently integrated into a perceptual grouping framework in which the visual tracking problem is tackled by identifying a subset of these line segments and connecting them sequentially to form a closed boundary with the largest saliency and a certain similarity to the previous one. Specifically, we define a new tracking criterion which combines a grouping cost and an area similarity constraint. The proposed criterion makes the resulting boundary tracking more robust to local minima. To achieve real-time tracking performance, we use Delaunay Triangulation to build a graph model with the detected line segments and then reduce the tracking problem to finding the optimal cycle in this graph. This is solved by our newly proposed closed boundary candidates searching algorithm called "Bidirectional Shortest Path (BDSP)". The efficiency and robustness of the proposed method are tested on real video sequences as well as during a robot arm pouring experiment.
		
\end{abstract}

	\section{INTRODUCTION}
	
\begin{figure}[tbp]
	\centering
	\subfloat{\includegraphics[width=1.5in]{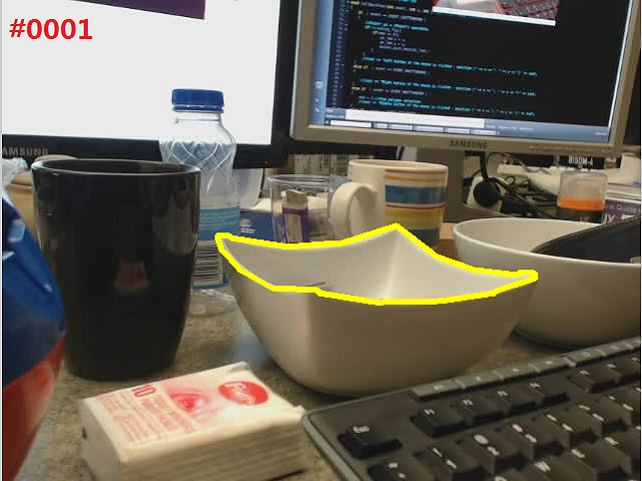}
		\label{0001}}
	\centering
	\subfloat{\includegraphics[width=1.5in]{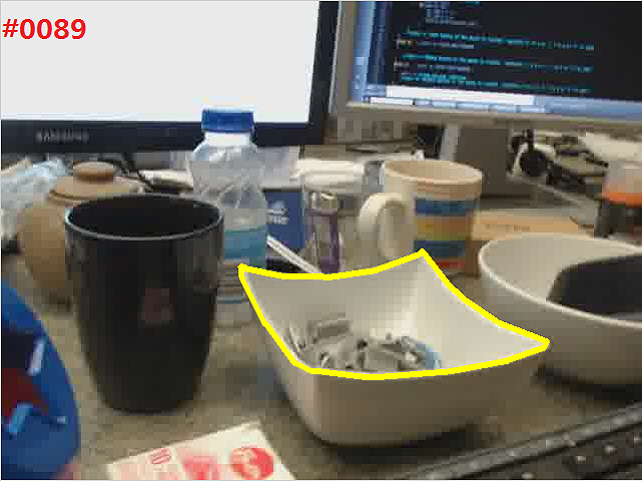}
			\label{0089}}\
		\centering
		\subfloat{\includegraphics[width=1.5in]{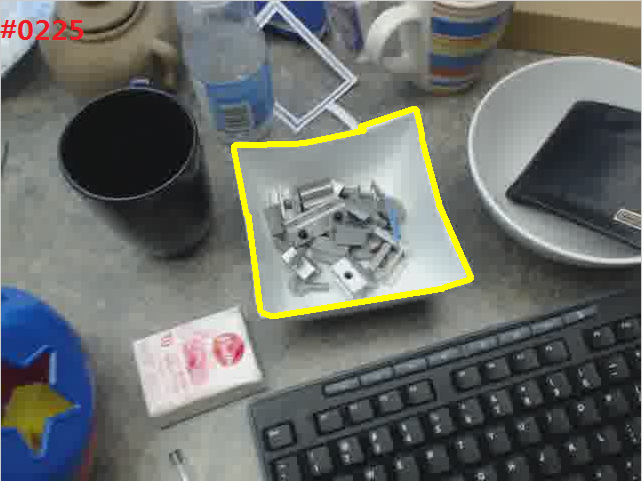}
			\label{0225}}
		\centering
		\subfloat{\includegraphics[width=1.5in]{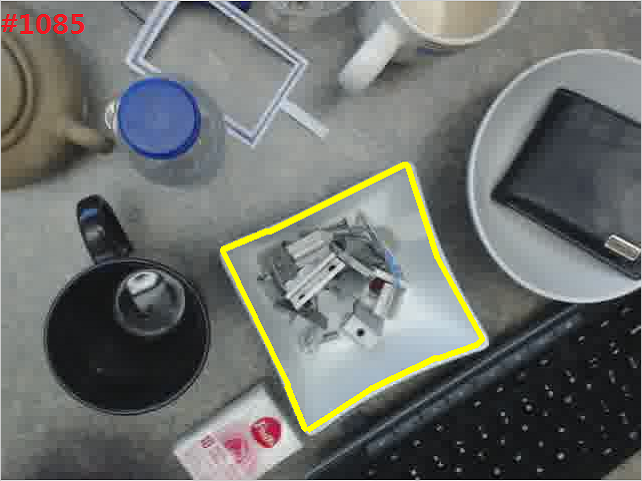}
			\label{1085}}
		\caption{Tracking the rim boundary of a bowl with the  following characteristics: (1) the rim of the bowl is non-planar, (2) the bowl itself has no salient stable textures, (3) the viewpoint changes dramatically.}
		\label{figNonplanarBowlExa}
\end{figure}

	Closed boundaries are common elements in real world scenes. Hence, real-time  closed boundary tracking is important in robot vision. As an example, consider Fig. \ref{figNonplanarBowlExa}, where the rim contour of a bowl is tracked. In real-world indoor images and robot applications, conventional trackers can fail, e.g. in the presence of texture-less target regions, non-rigid deformations, non-Lambertian surfaces, changing lighting conditions, cluttered background and drastic changing of supportive regions. 
	
  Template-based high DOF trackers can estimate image transformations such as affine and homography of a planar object region (or contour) from one frame to the next \cite{yilmaz2004object}. Many template trackers using different appearance models \cite{DBLP:conf/ijcai/LucasK81}, 
\cite{DBLP:conf/eccv/ScandaroliMR12},
\cite{DBLP:journals/pami/DowsonB08}
and search methods \cite{DBLP:journals/ijcv/BakerM04}, \cite{DBLP:conf/iros/BenhimaneM04}
have been proposed and achieve good performance. However, they are highly dependent on stable textures and sensitive to the presence of local minima. Keypoints based approaches \cite{DBLP:conf/cvpr/ShiT94}, 
\cite{Gauglitz_IJCV2011} are relatively robust to local minima, but they require many accurate feature points to be detected that can be difficult to achieve in practice.
  
    \rrr
    \kkk
	
  Non-planar contours tracking can be approached as a pose estimation problem 
\cite{DBLP:conf/iros/ChoiC12a} when 3D models are available. In unstructured, natural environments 3D models are seldom available. In these cases non-planar contours tracking can be approached as non-rigid tracking. Pixel-wise segmentation based \cite{DBLP:journals/cviu/GodecRB13} and contour evolving based 
\cite{DBLP:journals/pami/RathiVTY07},
\cite{DBLP:conf/cvpr/BibbyR10},
\cite{DBLP:journals/tip/0003YZL15} methods are usually employed to track those objects.
However, segmentation based methods do not work well in tracking targets whose appearances change significantly or targets which are comprised of several regions with great differences.
Given an initial contour from the previous frame, contour based tracking is performed by searching the target contour based on minimizing a suitable energy \cite{yilmaz2004object}. These methods are more likely to trap in local minima in cluttered environments \cite{DBLP:conf/iros/PressigoutM05}.
	
\begin{figure*}[htbp]
	\centering
	\subfloat[]{\includegraphics[width=.20\linewidth]{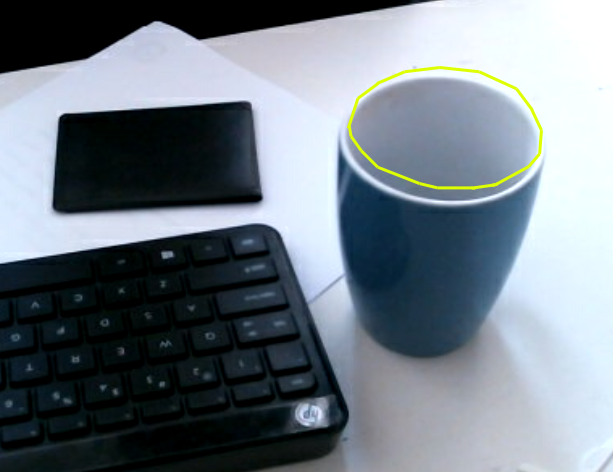}%
		\label{TrackingInput}}
	\hfil
	\centering
	\subfloat[]{\includegraphics[width=.20\linewidth]{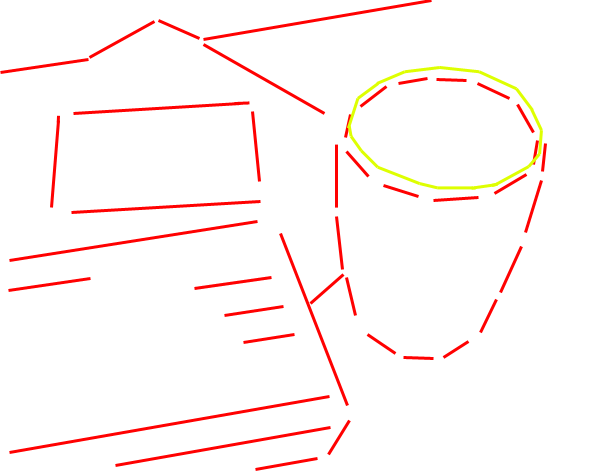}
		\label{LineSegments}}
	\hfil
	\centering
	\subfloat[]{\includegraphics[width=.20\linewidth]{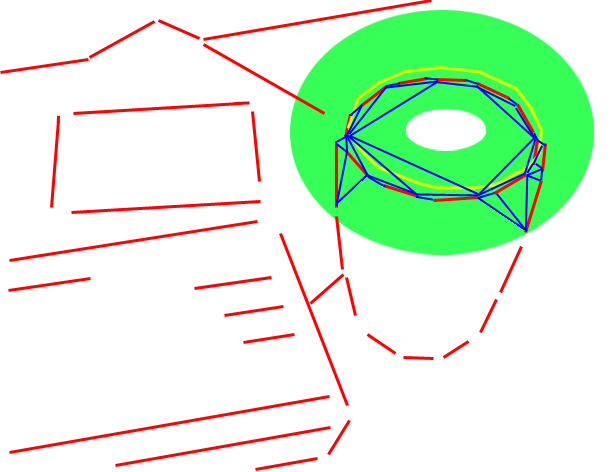}
		\label{GapFilling}}
	\hfil
	\centering
	\subfloat[]{\includegraphics[width=.20\linewidth]{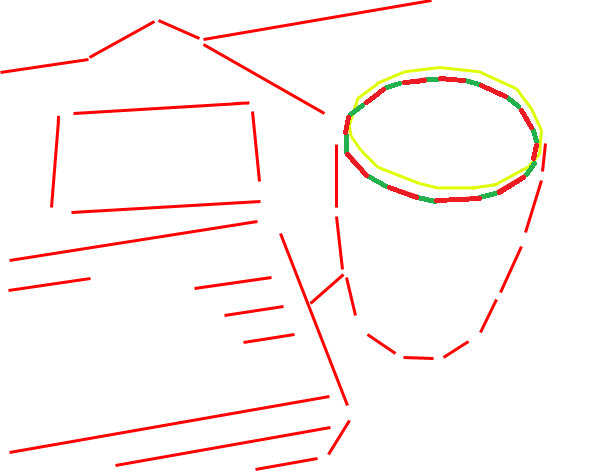}
		\label{TrackedBoudnary}}
	\caption{Illustration of closed boundary tracking: (a) Current image and the tracked boundary $(B_p)$ from the last frame (yellow polygon), (b) Detected line segments, (c) Gap filling among line segments in the buffer (green) region of boundary $B_p$, (d) Tracked boundary ($B_c$) of the current frame.}
	\label{figBoundaryTracking}
\end{figure*}

Another promising technique for closed boundary tracking is perceptual grouping, which has been widely used in salient closed boundary extraction from static images \cite{DBLP:conf/eccv/ElderZ96}, \cite{DBLP:journals/pami/WangKSW05}, \cite{DBLP:journals/tip/StahlW07}.
Schoenemann and Cremers \cite{DBLP:journals/pami/SchoenemannC10} perform contour tracking by integrating pixel-wise perceptual grouping and elastic shape matching into one solvable optimization problem. Unfortunately, this method is not real-time without using GPU processing, ruling out light weight or low power robotics applications. In \cite{DBLP:journals/ijon/LiuSFH12}, rough contour tracking and shape context matching are performed separately to improve time efficiency and handle cluttered background. However, the use of a fixed shape template restricts the ability to track non-planar closed boundaries with out-of-plane motion. 

Despite all different attempts, tracking of boundary targets in unstructured environment is still a challenging problem. 
This paper address this problem by presenting a novel line segments grouping based method for tracking salient closed boundaries. Our key contributions are:
\begin{itemize}
\item We define a salient closed boundary tracking criterion by combining a boundary grouping cost \cite{DBLP:journals/tip/StahlW07} and a regularization constraint on the boundary's area variation, which improves the tracking robustness greatly.
	 	
\item We construct a graph model $G(V,E)$ with the detected line segments by Delaunay Triangulation and develop a novel real-time searching method called Bidirectional Shortest Path searching algorithm (BDSP) for optimal closed boundary searching.
	 	
\item To evaluate the performance of our tracking scheme, we collected and annotated nine video sequences (9598 frames) of typical closed boundaries, which are challenging to track in real robot applications.
	 	
\end{itemize} 
	
  We implement our tracking method\footnote[1]{https://github.com/NathanUA/SalientClosedBoundaryTracking} and test it on our newly collected real world video sequences\footnote[2]{https://github.com/NathanUA/SalientClosedBoundaryTrackingDataset}, and compare it against state-of-the-art trackers: RKLT \cite{DBLP:conf/crv/ZhangSJ15}, ESM \cite{DBLP:conf/iros/BenhimaneM04}, HoughTrack \cite{DBLP:journals/cviu/GodecRB13} and a tracker adapted from the contour grouping method RRC \cite{DBLP:journals/tip/StahlW07}. We also test the performance of our method during a real robot arm experiment following a moving bowl and pouring cereal into it.
	
  The remainder of this paper is organized as follows. Section II formulates the problem by introducing the newly proposed tracking criterion.
Section III presents the details of the graph modeling and the novel optimization algorithm.
Section IV describes experimental results on the newly built real world dataset and an application of robot arm pouring. A brief conclusion is given in Section V.

\section{PROBLEM FORMULATION}
	
  Given a video sequence, we refer to the process of extracting corresponding salient closed boundaries from sequential video frames as boundary tracking. The target closed boundary is usually initialized by selecting a coarse polygon manually in the first frame. The main idea of the coming frames' boundary tracking is identifying a subset of line segments produced by line detector and connecting them to form a closed boundary which corresponds to the boundary tracked in the previous frame, see Fig. \ref{figBoundaryTracking}.

Therefore, the closed boundary tracking problem can be reduced to a prior shape constrained perceptual grouping problem. We define a tracking criterion which takes both grouping cost and shape constraint into consideration. The grouping cost $\Gamma$,  introduced in \cite{DBLP:journals/tip/StahlW07}, is given by:
\begin{equation}
\Gamma_{B} = \frac{|B_G|}{\iint_{R(B)}dx \,dy},
\label{Equ_cost}
\end{equation}
where $|B_G|$ denotes the summation of the gap filling segments (blue segments in Fig.\ref{GapFilling}) length along the boundary $B$. The denominator $\iint_{R(B)}dx \, dy$ is the area of region $R(B)$ which is enclosed by the boundary $B$.
	
\begin{figure}[tbp]
	\centering
	\subfloat[Lines within buffer region of prior boundary]{\includegraphics[width=0.9in]{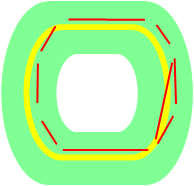}%
	\label{LinesPrior}}
	\hfill
	\subfloat[Wrong grouping]{\includegraphics[width=0.9in]{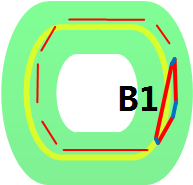}
		\label{WrongBoundary}}
	\hfill
	\subfloat[Correct grouping]{\includegraphics[width=0.9in]{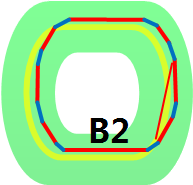}
		\label{CorrectBoundary}}
	\caption{Illustration of wrong boundary grouping: Yellow contours are prior boundaries. Line segments noise often results in wrong grouping of boundary ($B_1$) without using area constraint.}
	\label{figPriorConfusion}
\end{figure}

  Although the distance based filtering (the green region shown in Fig. \ref{GapFilling}) excludes many unrelated line segments, the above grouping cost still can not handle grouping illusions caused by noisy segments. As shown in Fig. \ref{figPriorConfusion}, it is clear that $\Gamma_{B_1} < \Gamma_{B_2}$ which results in an incorrectly grouped boundary $B_1$. To eliminate this kind of error without significantly increasing time complexity, we propose to use a simple area similarity $S_{Bp\_Bc} < S_e$ (e.g. 0.9) between the searched boundary ($B_c$) and the prior shape ($B_p$) to constrain the grouping process as follows:
\begin{equation}
	\centering
    S_{Bp\_Bc} = min(\frac{\iint_{R(Bp)}dxdy}{\iint_{R(Bc)}dxdy}, \frac{\iint_{R(Bc)}dxdy}{\iint_{R(Bp)}dxdy}),
	\label{EquAreaSimilarity}
\end{equation}
where $\iint_{R(Bp)}dxdy$ is the area of region $R(Bp)$ which is enclosed by the prior shape boundary $Bp$. $Bp$ is the initialization or the tracked boundary from the last frame. ($Bc$ is the to-be-tracked boundary). 

	
  To solve the grouping problem, we map the detected and generated (fill-in) line segments and their endpoints to an undirected graph $G = (V, E)$. Line segments and endpoints correspond to graph edges and graph vertices respectively, and thus closed boundaries correspond to graph cycles. Now, the problem of closed boundary grouping is converted into an optimal graph cycle searching problem. We develop a novel graph based optimization method to find the optimal boundary (quasi-optimum) in real-time.
	
\section{METHOD}
	
\begin{figure}[tbp]
	\centering
	\includegraphics[width=2.4 in]{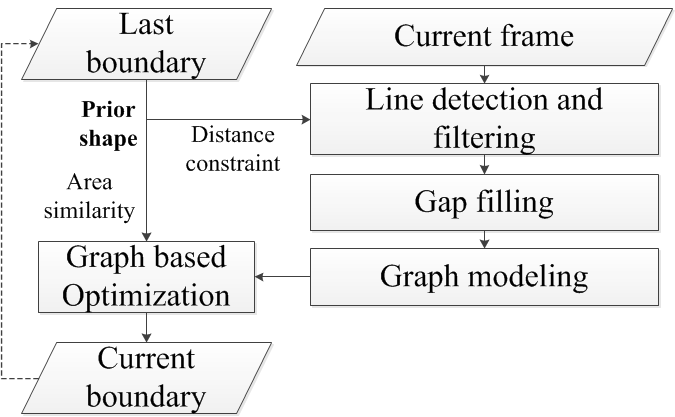}
	\caption{Workflow of boundary tracking}
	\label{figWorkflow}
\end{figure}

  The workflow of our salient closed boundary tracking method is shown in Fig. \ref{figWorkflow}.
First, we  introduce the line detection and filtering. Then, the details of gap filling are presented, followed by graph modeling and optimization.
	
\subsection{Line Detection and Filtering}
	
  Straight line segments are fundamental elements in our tracking method. 
  We use EDLines \cite{DBLP:journals/prl/AkinlarT11}, a real-time line segments detector, for automatic detection of boundary line segments. Detected line segments are represented by pairs of endpoints.
In each incoming frame, line detection is conducted on the whole frame. Lines of interest are filtered by a distance constraint (with similar effect of the green buffer region shown in Fig. \ref{GapFilling}) from the previous boundary. Specifically, three distances of a line from its two end-points and mid-point to the prior boundary are computed. If the average of these distances is smaller than certain threshold (e.g. 20 pixels), it will be retained.
Lines of interest are not directly detected from the frame subregion defined by the green buffer because masking an image with an irregular buffer region takes more time.
    
	

	
\subsection{Gap Filling}
\begin{figure*}[htbp]
	\centering
	\subfloat[Line segments]{\includegraphics[height=0.8in]{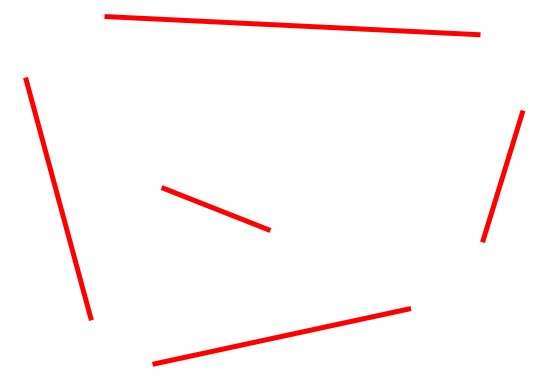}%
		\label{lineSegments_dt}}
		\hfill
	\centering
	\subfloat[Endpoints]{\includegraphics[height=0.8in]{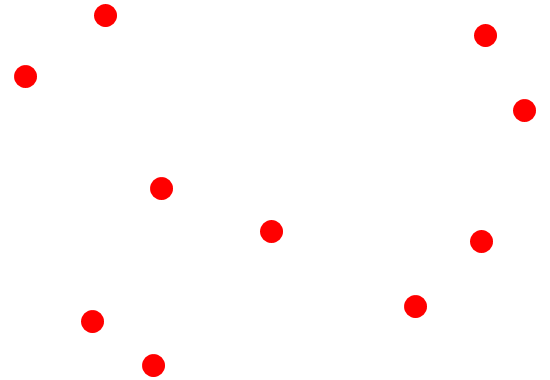}
		\label{endpoints_dt}}
		\hfill
	\centering
	\subfloat[Gap filling by DT]{\includegraphics[height=0.8in]{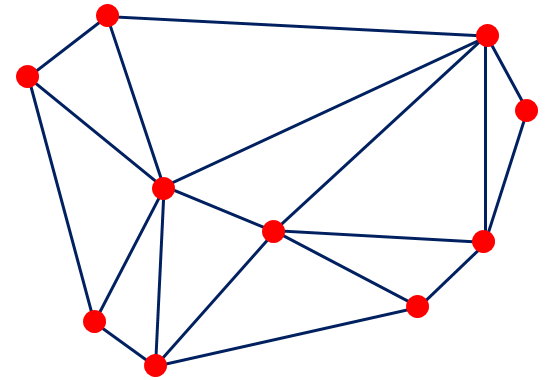}
		\label{gapFilling_dt}}
		\hfill
	\centering
	\subfloat[Graph structure]{\includegraphics[height=0.8in]{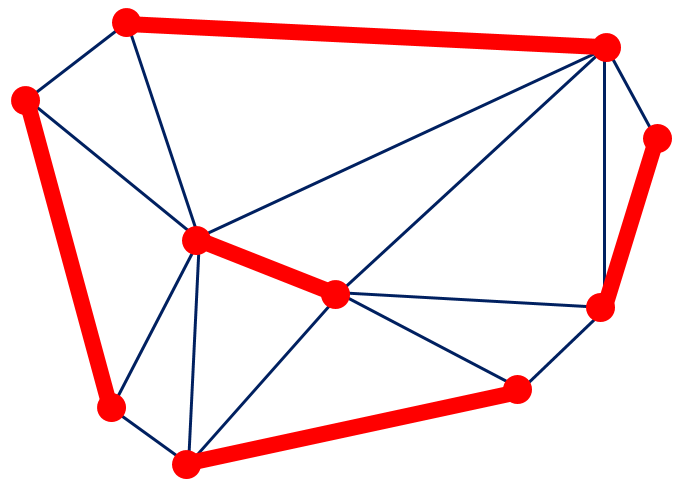}
		\label{finalGapFilling}}
        \hfill
	\centering
	\subfloat[Cycle searching]{\includegraphics[height=1in]{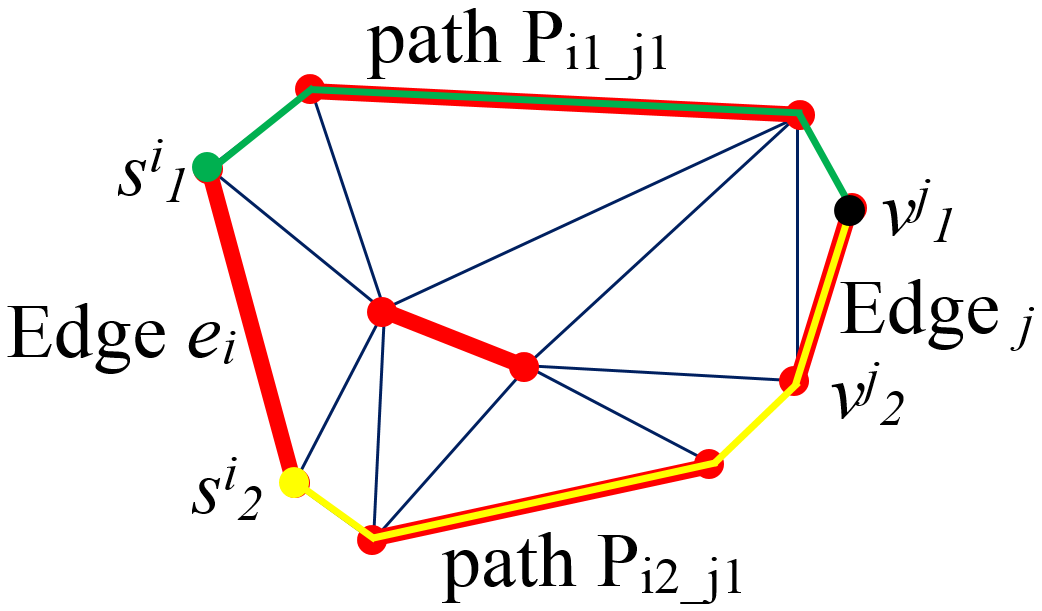}
		\label{BDSP}}
	\caption{(a)-(d) Graph structure construction by gap filling, (e) Cycle candidates searching by BDSP: Edge $e_i$ is the current edge and $v^j_1$ is the third vertex. Edge $e_i$, shortest paths $P_{i1\_j1}$ and path $P_{i2\_j1}$ construct a closed cycle candidate.}
	\label{figGapFillingProcess}
\end{figure*}

	After obtaining the line segments of interest, Delaunay Triangulation (DT) \cite{DBLP:books/sp/PreparataS85} is introduced to generate virtual fragments and fill the gaps among disconnected segments, as illustrated in Fig. \ref{figGapFillingProcess}. First, line segments are simplified as endpoints (see Fig. \ref{lineSegments_dt} and Fig. \ref{endpoints_dt}). 
	Then, DT is conducted on these endpoints (Fig. \ref{gapFilling_dt}).
	Finally, we superimpose the detected line segments in Fig. \ref{lineSegments_dt} over the generated DT edges in Fig. \ref{gapFilling_dt}.
	Generated DT edges that overlap the detected line segments are removed. The final result of gap filling is an undirected graph structure as shown in Fig. \ref{finalGapFilling}. To distinguish two kinds of line segments, we refer to the detected line segments (the red lines in Fig. \ref{finalGapFilling}) as \emph{detected} segments and the gap filling segments (the blue lines in Fig. \ref{finalGapFilling}) as \emph{generated} segments.
    
\subsection{Graph Modeling}
	The gap filling process constructs the structure of an undirected graph $G = (V, E)$, which maps endpoints and segments (both \emph{detected} and \emph{generated}) to graph vertices $V$ and edges $E$ respectively. Then, we define the edge-weight function for each edge $e \in E$ similar to \cite{DBLP:journals/tip/StahlW07}. Given a graph edge $e_i$, we set its weight to
\begin{equation}
	w(e_i) = \left\{
	\begin{aligned}
	0   & &\text{$e_i$ is a \emph{detected} segment}\\
	|P^i_1P^i_2|   & &\text{$e_i$ is a \emph{generated} segment}\\
	\end{aligned}
	\right.
\end{equation}
where $|P^i_1P^i_2|$ is the length of the corresponding line segment $P^i_1P^i_2$ of the graph edge $e_i$.
	
\subsection{Graph based Optimization}
	
  Now, our goal is to find the optimal graph cycle which has the minimum boundary cost, based on (\ref{Equ_cost}), and satisfies the similarity constraint in (\ref{EquAreaSimilarity}) simultaneously. The key problem is that both the boundary cost and the similarity constraint cannot be determined when the boundary itself is unknown. Furthermore, they are difficult to be integrated into one cost function. So we developed a novel heuristic search method to obtain the optimal cycle. Our method has two steps: generating boundary candidates and finding the optimal one from these candidates.
	
 Given a graph, exhaustive searching is time-consuming and unfavorable. In an attempt to avoid an exhaustive search, we propose a method called ``Bidirectional Shortest Path (BDSP)'' to generate cycle candidates. This method is based on the hypothesis that a cycle with smaller total weight is more likely to be the optimal boundary. Given a graph which contains $n$ \emph{detected} line segments, we sample half $n/2$ of them and search $(n-1)$ (the current segment is excluded) cycle candidates for each sampled (\emph{detected}) segment. As shown in Fig. \ref{BDSP}, for each sampled \emph{detected} edge $e_i$(edges with odd or even indices), we search shortest paths from its two vertices $s^i_1$ and $s^i_2$ to the same third vertex $v^j_1$ using Dijkstra \cite{dijkstra1959note}. The weight of $e_i$ is set to infinity other than its original weight zero during searching. The edge $e_i$, shortest paths $P_{i1\_j1}$ and $P_{i2\_j1}$ construct a cycle candidate $C_{i\_j1}$. Vertex $v^j_1$ and Vertex $v^j_2$ belong to the same edge $e_j$ and the weight of $e_j$ is zero, thus, taking $v^j_1$ or $v^j_2$ as the third vertex usually produce the same cycle.  Hence, the total number of the cycle candidates is $n (n-1)/2$. 
Having obtained these cycle candidates, we can search for the optimal closed boundary by computing their boundary costs (\ref{Equ_cost}) and similarity constraints (\ref{EquAreaSimilarity}) easily. The optimal boundary searching algorithm is shown in Algorithm  \ref{algoBoundarySearching}. 
	
\begin{algorithm}[tbp]
	\caption{Optimal cycle searching} 
	\hspace*{0.02in} {\bf Input:}
	Undirected graph $G = (V,E)$ and a shape prior represented by a set of ordered points\\
	\hspace*{0.02in} {\bf Output:}
	The optimal cycle $C_{opt}$
	\begin{algorithmic}[1]
		\For{$i=0;i<n;i+=2$}
		\State Use BDSP to search cycle candidates $C_{i\bullet}$
		\For{$j=0;j<2(n-1);j+=2$}
		\If{$i==0\&\&j==0$}
		\State$C_{opt} = C_{ij}$
		\EndIf
		\If{$S_{C_{ij}\_Bp}$ > $S_e$}
		\If{$\Gamma_{C_{ij}} < \Gamma_{C_{opt}}$}
		\State $C_{opt} = C_{ij}$
		\EndIf		
		\EndIf
		\EndFor
		\EndFor
		\State \Return $C_{opt}$
	\end{algorithmic}
Notes: $C_{i\bullet}$ are the $(n-1)$ cycle candidates related to edge $e_i$.
	\label{algoBoundarySearching}
\end{algorithm}
	
\section{EXPERIMENTAL RESULTS}
	
  We validate our tracking scheme using a number of comparative experiments and a real robot arm experiment.
	
	
\subsection{Dataset and Evaluation Measure}
	
  We collected nine video sequences of salient closed boundaries, as shown in Fig. \ref{figResults}. Each sequence is about 30 sec (30 fps) and the frame size is 640$\times$480 (width$\times$height). There are 9598 frames in total. In each sequence, different motion styles such as translation, rotation and viewpoint changing are all performed. We annotate them by drawing polygons which are well matched with the salient closed boundaries in human vision.
	
  To evaluate our proposed method quantitatively, we define the error metric as the alignment error ($E_{AL}$) of tracked closed boundary and ground truth ($B_{gt}$) as:
$
E_{AL} = 	max\{\frac{B_i\otimes Dist_{B_{gt}}}{P_{B_i}},\frac{B_{gt}\otimes Dist_{B_i}}{P_{B_{gt}}}\},
$
where $\otimes$ denotes convolution, 
$B_i$ is the boundary binary image, $Dist_{B_i}$ is the distance transform image of $B_i$ and $P_{B_i}$ indicates the perimeter of boundary $B_i$.
We use the success rate to measure a tracker's overall accuracy \cite{DBLP:conf/crv/SinghRZJ16}. The success rate on a sequence is defined as the ratio of frames where the tracking error $E_{AL}$ is less than a threshold of $e_p$ pixels and the total frames.	
	
\subsection{Results}
	
  We compare our tracking method BDSP against following methods: ESM \cite{DBLP:conf/iros/BenhimaneM04} which is a popular registration based homography tracker, RKLT \cite{DBLP:conf/crv/ZhangSJ15} which is a cascade registration based tracker that can handle partial appearance changing and occlusion by RANSAC, HoughTrack \cite{DBLP:journals/cviu/GodecRB13} which is a state-of-the-art segmentation based tracker that can provide us accurate contour and a tracker adapted from edge grouping method RRC \cite{DBLP:journals/tip/StahlW07}. Both ESM and RKLT are tested with appearance model NCC, which are implemented in a modular tracking framework (MTF) \cite{DBLP:conf/crv/SinghRZJ16}). We initialize them by selecting a quadrilateral which encloses the target boundary at the first frame. Then boundaries of following frames are computed by homography transformations with respect to the first frame (all boundaries are assumed to be planar). We modified RRC by substituting its line detector for EDlines \cite{DBLP:journals/prl/AkinlarT11} and added a buffer search region as shown in Fig. \ref{GapFilling}. 

\begin{figure*}[htbp]
	\centering
	\subfloat[Bowl rim]{\includegraphics[width=0.191\linewidth]{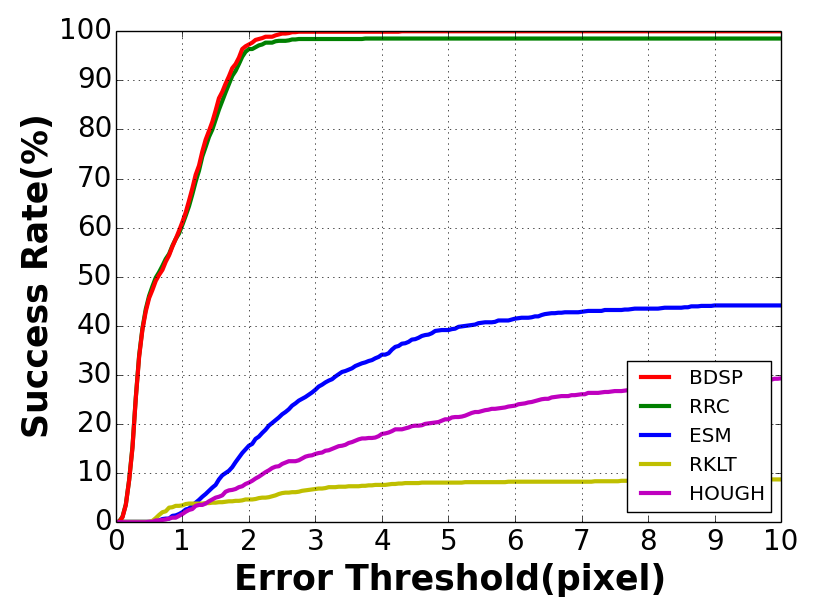}%
		\label{Bowl}}
	\centering
	\subfloat[Garbage bin rim]{\includegraphics[width=0.191\linewidth]{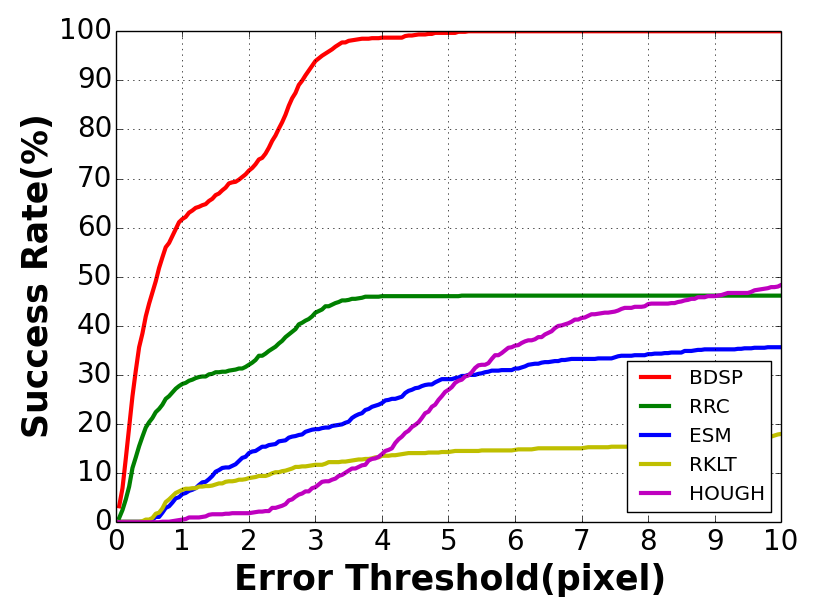}
		\label{GarbageBin}}
	\centering
	\subfloat[Transparent cup rim]{\includegraphics[width=0.191\linewidth]{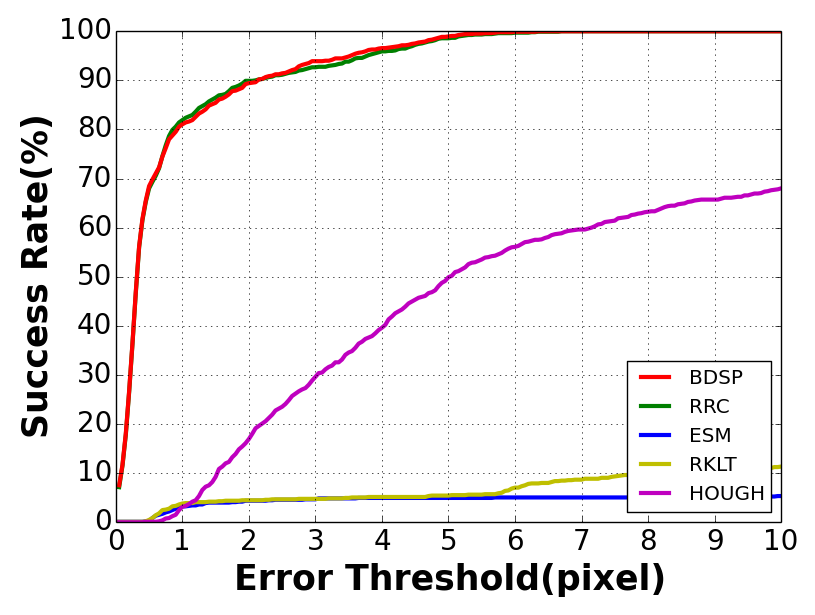}
		\label{TransparentCup}}
	\centering
	\subfloat[Mark cup rim]{\includegraphics[width=0.191\linewidth]{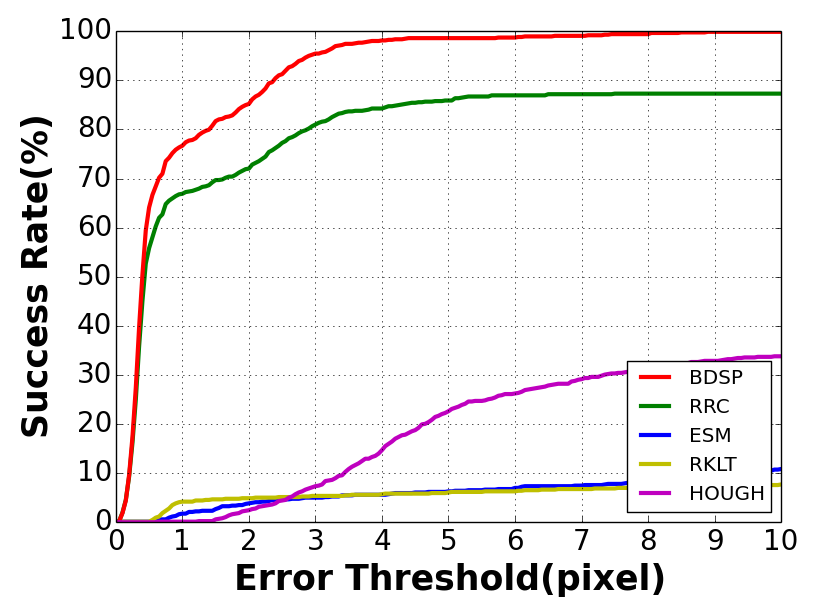}
		\label{MarkCup}}\\
	\centering
	\subfloat[Mark cup pouring tea]{\includegraphics[width=0.191\linewidth]{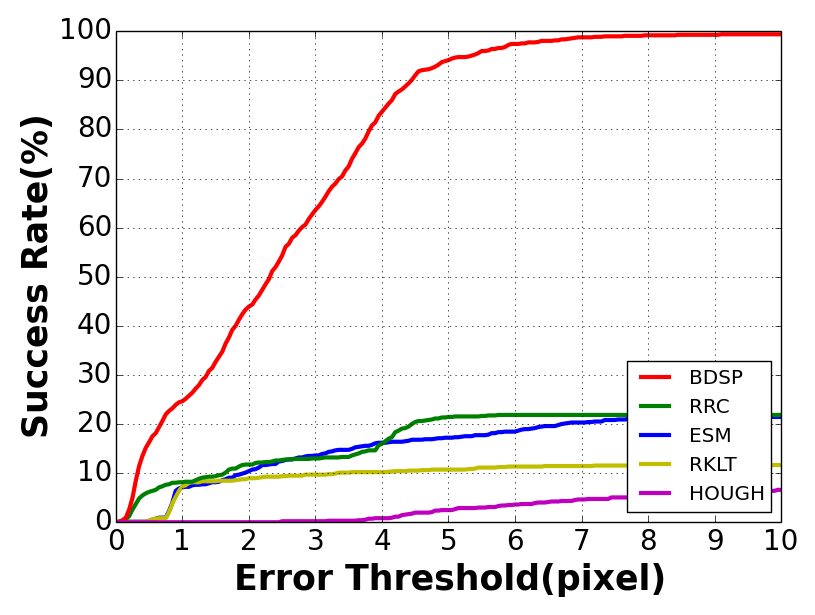}
		\label{MarkCupPourWater}}
	\centering
	\subfloat[Tool box rim]{\includegraphics[width=0.191\linewidth]{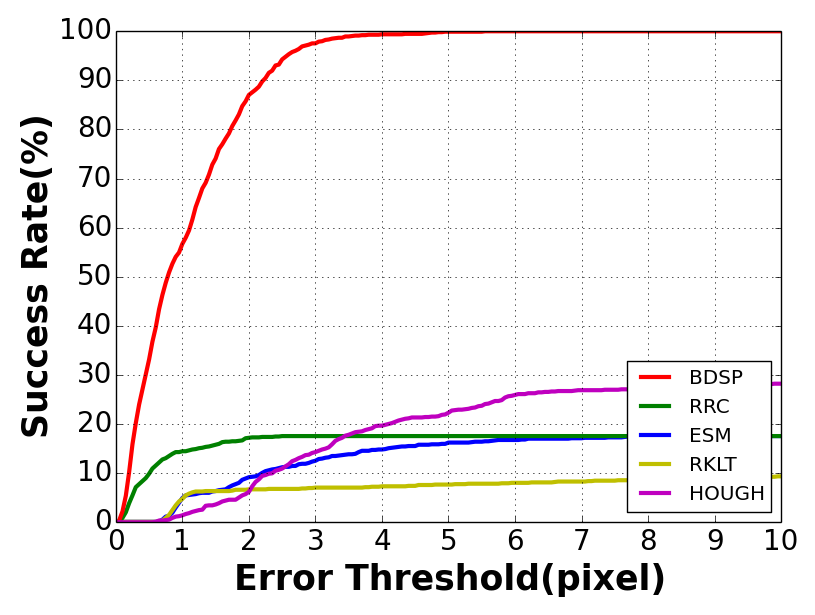}
		\label{ToolBox}}
	\centering
	\subfloat[Non-planar bowl rim]{\includegraphics[width=0.191\linewidth]{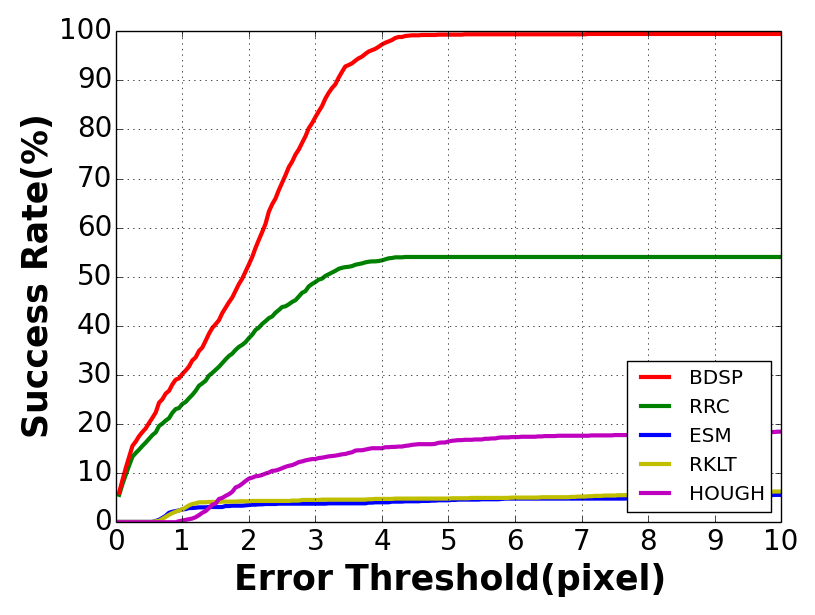}
		\label{NonplanarBowl}}
	\centering
	\subfloat[Mark cup contour]{\includegraphics[width=0.191\linewidth]{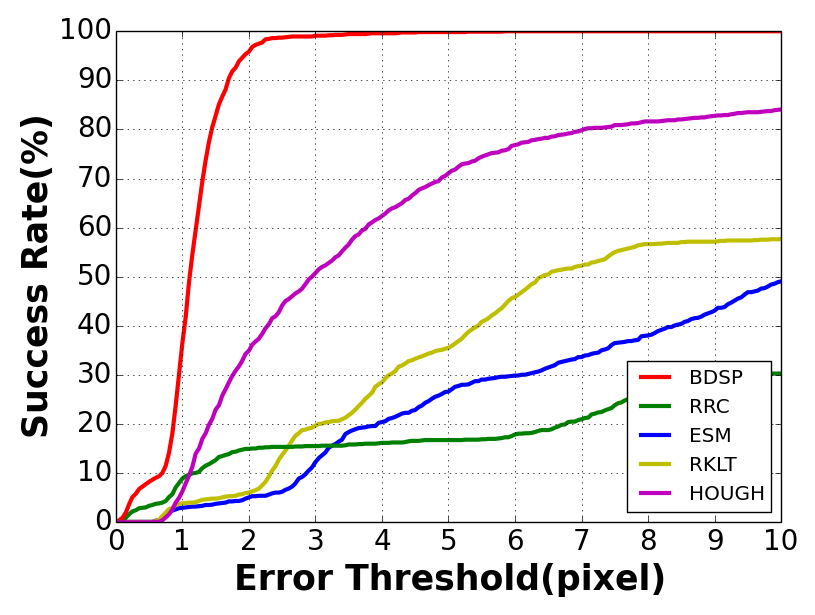}
		\label{MarkCupShape}}
	\centering
	\subfloat[Book stand contour]{\includegraphics[width=0.191\linewidth]{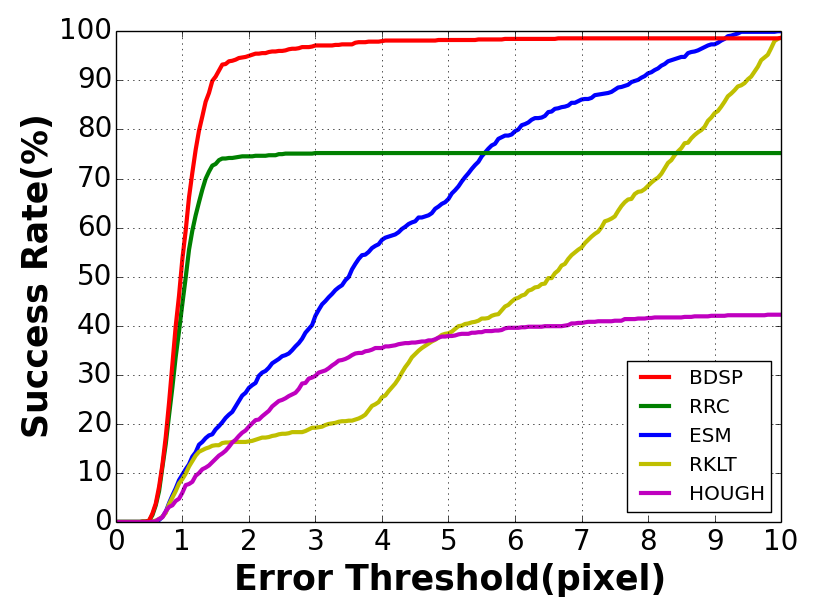}
		\label{BookStand}}
	\caption{Success rates of BDSP, RRC, ESM, RKLT and HoughTrack on the video sequences of Fig. \ref{figResults}.}
	\label{figSuccessRate}
\end{figure*}
\begin{figure*}
	\centering
	\subfloat[Bowl rim]{\includegraphics[width=0.19\linewidth]{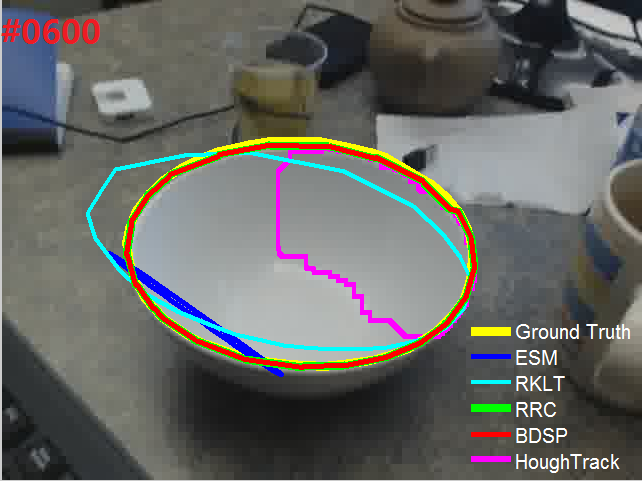}%
		\label{Bowl600rlt}}
        \hspace{0.0001in}
	\centering
	\subfloat[Garbage bin rim]{\includegraphics[width=0.19\linewidth]{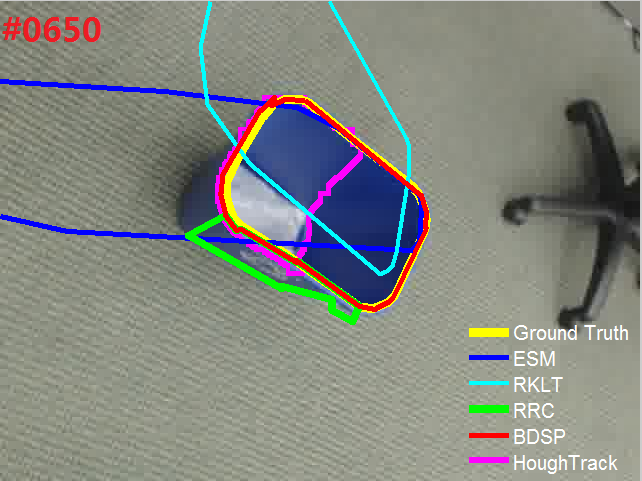}%
		\label{GarbageBin650rlt}}
        \hspace{0.0001in}
	\centering
	\subfloat[Transparent cup rim]{\includegraphics[width=0.19\linewidth]{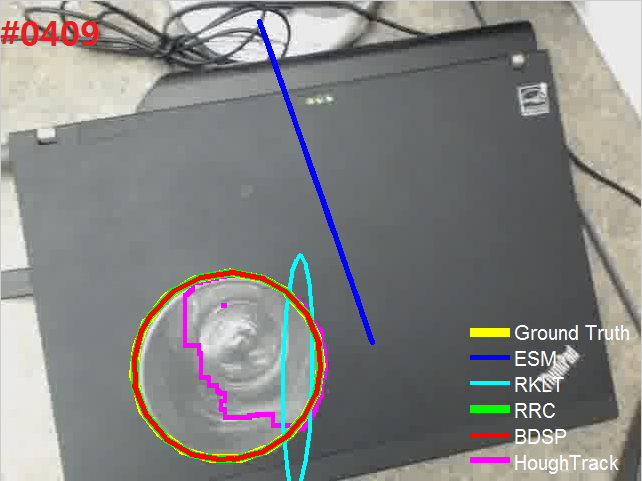}%
		\label{TransparentCup409rlt}}
        \hspace{0.0001in}
	\centering
	\subfloat[Mark Cup rim]{\includegraphics[width=0.19\linewidth]{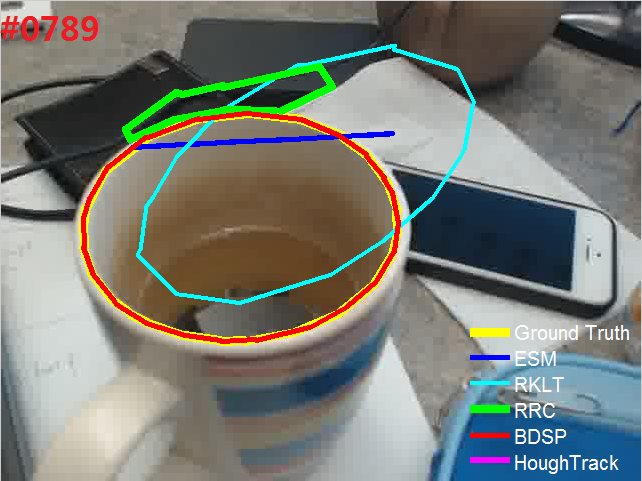}%
		\label{MarkCup789rlt}}\\
	\centering
	\subfloat[Mark cup rim pouring]{\includegraphics[width=0.19\linewidth]{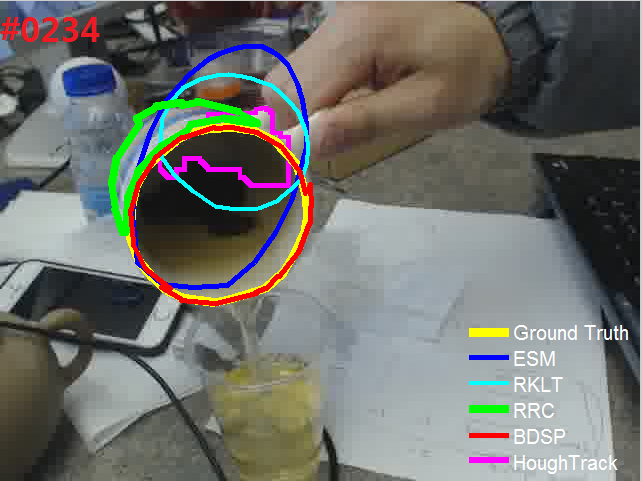}%
		\label{MarkCupPourWater234rlt}}
        \hspace{0.0001in}
	\centering
	\subfloat[Tool box rim]{\includegraphics[width=0.19\linewidth]{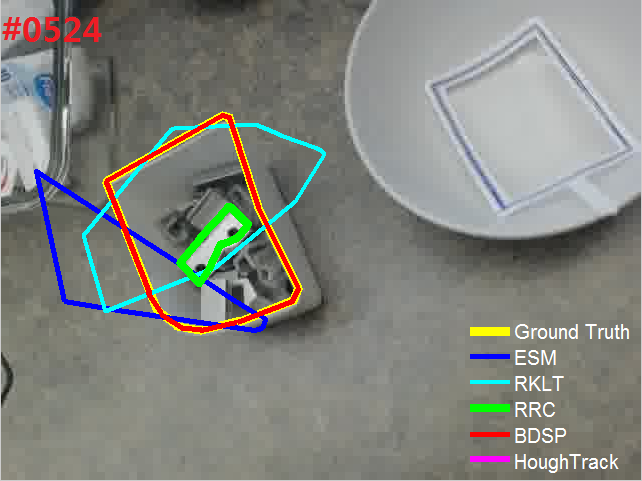}%
		\label{ToolBox524rlt}}
        \hspace{0.0001in}
	\centering
	\subfloat[Non-planar bowl rim]{\includegraphics[width=0.19\linewidth]{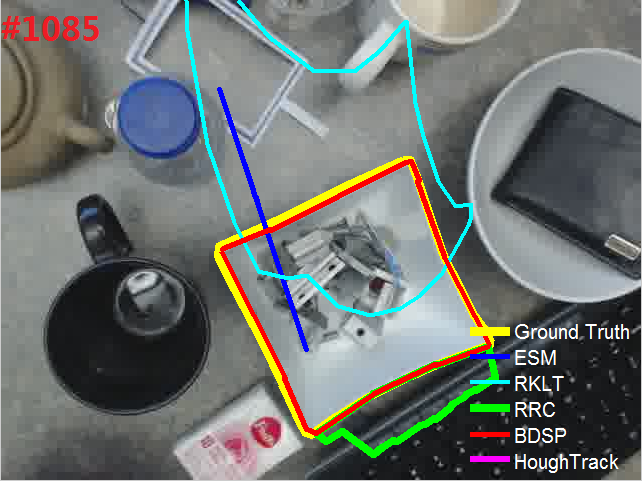}%
		\label{Nonplanar1085rlt}}
		\hspace{0.0001in}
	\centering
	\subfloat[Mark cup contour]{\includegraphics[width=0.19\linewidth]{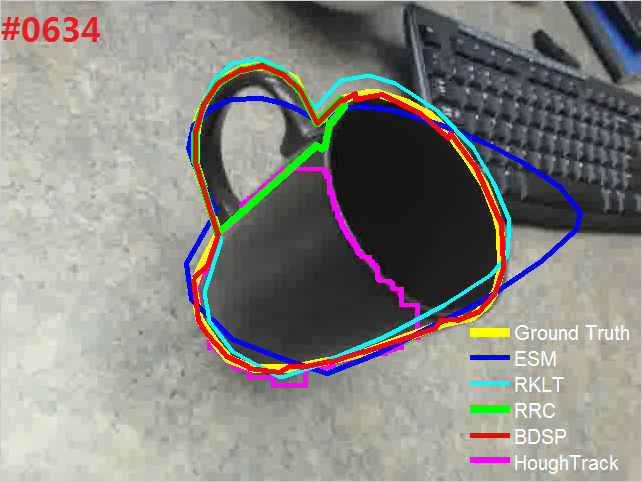}%
		\label{MarkCupShape634rlt}}
        \hspace{0.0001in}
	\centering
	\subfloat[Book stand contour]{\includegraphics[width=0.19\linewidth]{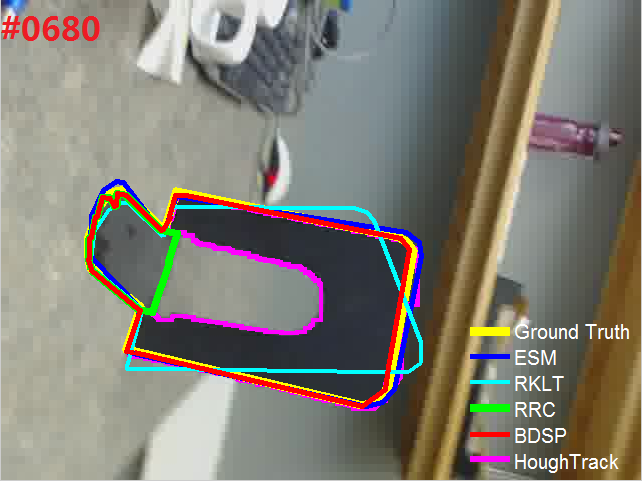}%
		\label{BookStand680rlt}}
	\caption{Tracked boundaries on typical frames.}
	\label{figResults}
\end{figure*}

  The success rate curves of the above four methods and our method are illustrated in Fig. \ref{figSuccessRate}. As we can see, the proposed method (BDSP) performs better than others in almost all of these sequences. Both registration based trackers ESM and RKLT fail quickly because they are sensitive to appearance changing, as blue and cyan boundaries shown in Fig. \ref{Bowl600rlt} to Fig. \ref{Nonplanar1085rlt}. Although HoughTrack performs better than registration based trackers thanks to its model updating, it corrupts quickly when the content of target region are heterogeneous as pink boundaries shown in Fig. \ref{figResults}. Fig. \ref{Bowl600rlt} and Fig. \ref{TransparentCup409rlt} show an empty bowl and an empty transparent cup with relative clean background. The corresponding tracking results illustrated in Fig. \ref{Bowl} and Fig. \ref{TransparentCup} show that the RRC tracker produces almost the same success rate with our method. But it is very susceptible to noise, as the tracked green boundaries shown in Fig. \ref{figResults}. The intact video results are included in the supplementary video\footnote[3]{https://youtu.be/RXjD0yHkukI}.

\begin{table}[tbp]
	\centering
    \caption{GRAPH SCALE AND TIME EFFICIENCY}
	\begin{tabular}{p{0.2in}p{0.17in}p{0.17in}p{0.17in}p{0.17in}p{0.17in}p{0.17in}p{0.17in}p{0.17in}p{0.17in}}
		\toprule
		Video & Fig.\ref{Bowl600rlt} & Fig.\ref{GarbageBin650rlt} & Fig.\ref{TransparentCup409rlt} & Fig.\ref{MarkCup789rlt} & Fig.\ref{MarkCupPourWater234rlt} & Fig.\ref{ToolBox524rlt} & Fig.\ref{Nonplanar1085rlt} & Fig.\ref{MarkCupShape634rlt} & Fig.\ref{BookStand680rlt} \\
		\midrule
		Edges & 445 & 508 & 550 & 672 & 753 & 362 & 419 & 505 & 286\\
		Nodes & 80 & 88 & 96 & 112 & 124 & 62 & 70 & 89 & 53\\
		$lt$(ms) & 4.34 & 8.78 & 4.15 & 4.69 & 5.17 & 4.33 & 5.09 & 5.07 & 3.79\\
		$gt$(ms) & 14.45 & 18.85 & 19.00 & 26.10 & 32.13 & 8.68 & 16.93 & 17.72 & 6.56\\
		fps & 54.12 & 37.55 & 43.21 & 32.48 & 26.80 & 76.81 & 59.74 & 43.88 & 96.57\\
		\bottomrule
	\end{tabular}
	\label{tabTimer}
    \\
   \leftline{Notes: ms denotes millisecond.}
\end{table}
	
  We also measured the average processing speed of our method for each of the nine video sequences on a machine with a quad core 3.10 GHz Intel Core i5 processor, 16 GB RAM and Ubuntu 14.04 64-bit OS. Our method is implemented in C++ using OpenCV and Boost library. Table. \ref{tabTimer} illustrates the average graph size, which is indicated by numbers of edges and nodes, the average time costs of line detection ($lt$), grouping time ($gt$) and the average frequency per second (fps). The total tracking time per each frame contains line detection time and grouping time. We compute the instantaneous fps of each frame and then average them over the whole sequence to get the average fps, as shown in the last row of Table \ref{tabTimer}.  As we can see our method is acceptable for real-time tracking.
	
\subsection{Robot Arm Pouring Experiment}
  Our salient closed boundary tracker has been used successfully in a real robot arm pouring experiment. The task is to track and follow a moving bowl and then pour cereal into it. The difficulties of this experiment are that the bowl is non-Lambertian and has no salient textures. 
	
\begin{figure*}[tbp]
	\centering
	\subfloat[Experiment setup]{\includegraphics[width=.24\linewidth]{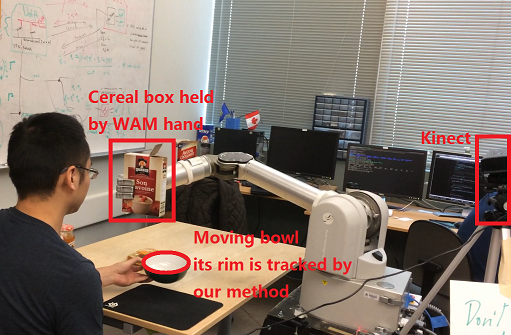}
		\label{pouringPhoneRecord2}}
	\centering
	\subfloat[Following]{\includegraphics[width=0.24\linewidth]{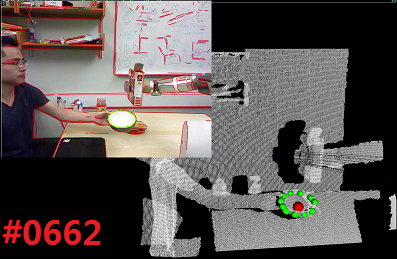}
		\label{robot_0662}}
	\centering
	\subfloat[Following]{\includegraphics[width=0.24\linewidth]{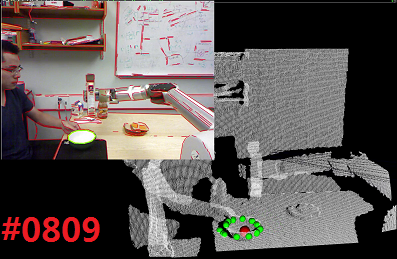}
		\label{robot_0809}}
	\centering
	\subfloat[Following and accurate pouring]{\includegraphics[width=.24\linewidth]{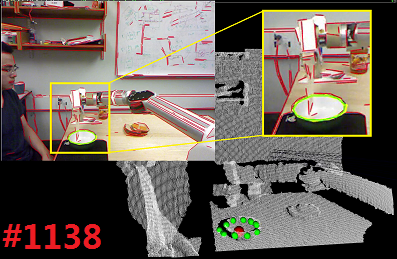}
		\label{robot_1138}}
	\caption{Robot pouring experiment: (a) A human is moving the bowl continuously under the surveillance of a Kinect. (b)(c) Our algorithm tracks the bowl rim and maps its 2D image points to 3D points in the robot coordinates through the Kinect which is registered with the robot coordinates, then, makes the robot arm follows the moving bowl. (d) The robot hand pours the cereal into the bowl accurately according to our tracking result.}
	\label{figRobotPouring}
\end{figure*}

  The setup of our experiment is shown in Fig. \ref{pouringPhoneRecord2}. The system includes a set of WAM arm and a Kinect. The 3D coordinates of the WAM arm and the Kinect are registered. We initialize the bowl rim and track it in RGB video stream captured by the Kinect. Meanwhile, we map the tracked closed boundary, which is represented by a set of 2D image points, to 3D points acquired by the Kinect depth sensor, as shown in Fig. \ref{robot_0662} and Fig. \ref{robot_0809}. The centroid of the bowl is computed based on these mapped 3D contour points and is taken as the pouring target position. We pre-compute the shifting of the WAM hand to the bowl centroid. When the distance $Dist_{r\_b}$ between the centroid of the tracked bowl and the robot hand satisfies certain thresholds($e_{low}<Dist_{r\_b}<e_{high}$), the WAM arm will pour the cereal into the bowl, as shown in Fig. \ref{robot_1138}. Without having the tracking points of the contour provided by our tracker it will be very difficult to find the 3D center of the bowl with any other types of trackers. The experiment shows that our tracker is stable and efficient in real robot application. A demonstration of the pouring task can be seen in the accompanying video.

\section{CONCLUSIONS}
	
	We presented a novel real-time method for salient closed boundary tracking. By combining a saliency measure and an area constraint as tracking criterion, the proposed method improves the tracking performance greatly. A bidirectional shortest path based boundary candidates searching algorithm enables the real-time solvability of the combined tracking criterion. We validated it quantitatively on various real-world video sequences. Since it is robust and fast enough, it has been used successfully in real robot pouring experiment where other trackers have failed.
    Our future work will focus on addressing the problem of tracking boundaries, which are hard to be described by straight line segments, as well as the problem of self-occlusions.
    
	\bibliographystyle{IEEEtran}
	\bibliography{IEEEabrv,BoundaryTracking}
	
\end{document}